\documentclass[12pt]{article}

\usepackage[T1]{fontenc}
\usepackage[active]{srcltx}
\usepackage{lmodern}
\usepackage{amsmath,amstext,amssymb,graphicx,graphics,color}
\usepackage{theorem}
\usepackage{cite}               
\usepackage{hyperref}           
\usepackage{bbm}                
\usepackage{fancybox}           

\theorembodyfont{\normalfont\slshape} 

\theoremstyle{break} 

{{\par\noindent \bf Proof. \nobreak}}%
{\nobreak \removelastskip \nobreak \hfill $\Box$ \medbreak}
{{\par\noindent \bf Proof \nobreak}}%
{\nobreak \removelastskip \nobreak \hfill $\Box$ \medbreak}
{{\par\noindent \bf Proof lemma. \nobreak}}%
{\nobreak \removelastskip \nobreak \bf End proof lemma. \medbreak}

\usepackage{fancyhdr}
 \fancypagestyle{plain}{
  \fancyhead{}
  
}

\def\paragraph#1{{\bf #1\ }}

\newcommand{\expo}{\mathrm{e}}

\usepackage[utf8]{inputenc}

\usepackage{authblk}

\title{Lesion segmentation using U-Net network}
\author[1]{Adrien Motsch}
\author[2]{Sébastien Motsch}
\author[3]{Thibaut Saguet}
\affil[1,3]{Virtual Expertise, Clinique pasteur, Toulouse (France)}
\affil[2]{Arizona State University, Tempe (USA)}

\begin{document}

\maketitle

\begin{abstract}
  This paper explains the method used in the segmentation challenge (Task 1) in the International Skin Imaging Collaboration's (ISIC) Skin
Lesion Analysis Towards Melanoma Detection challenge held in 2018. We have trained a U-Net network to perform the segmentation. The key elements for the training were first to adjust the loss function to incorporate unbalanced proportion of background and second to perform post-processing operation to adjust the contour of the prediction.
\end{abstract}

\section{Introduction}

\begin{figure}[ht]
  \centering
  \includegraphics[scale=.8]{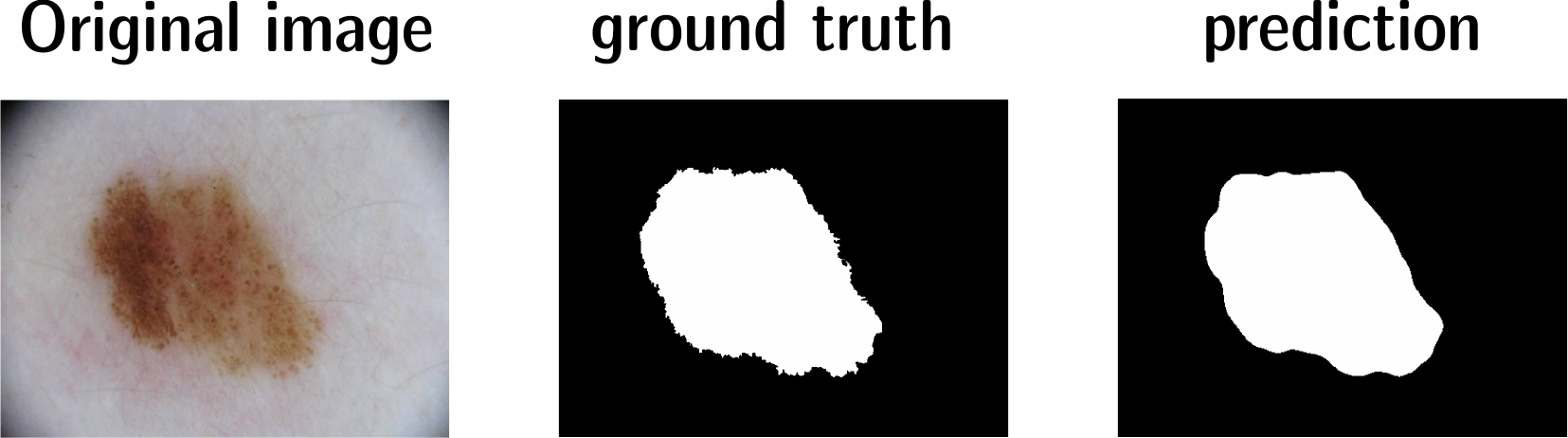}
  \caption{Example of segmentation (image \textit{ISIC\_0000009.jpg} from the training set 2018): the original image is on the left, in the middle the mask manually generated ('ground truth'), and on the right the prediction of our network based on the U-Net architecture.}
  \label{fig:example_segmentation}
\end{figure}

Segmentation is an important task to highlight zone of interests in medical images. In the ISIC 2018 challenge, the goal is to perform lesion boundary segmentation. To perform the segmentation, we have chosen to use a U-Net architecture \cite{ronneberger_u-net:_2015} as it is known to perform well in medical image and without a large data-set. The details of the training is explained below and an example of mask generation is given in figure \ref{fig:example_segmentation}. Our data was extracted from the \textit{ISIC 2018: Skin Lesion Analysis Towards Melanoma Detection} grand challenge data-sets \cite{valle_data_2017,tschandl_ham10000_2018}.


\section{Methods}

\subsection{Pre-processing}

\begin{figure}[ht]
  \centering
  \includegraphics[scale=.45]{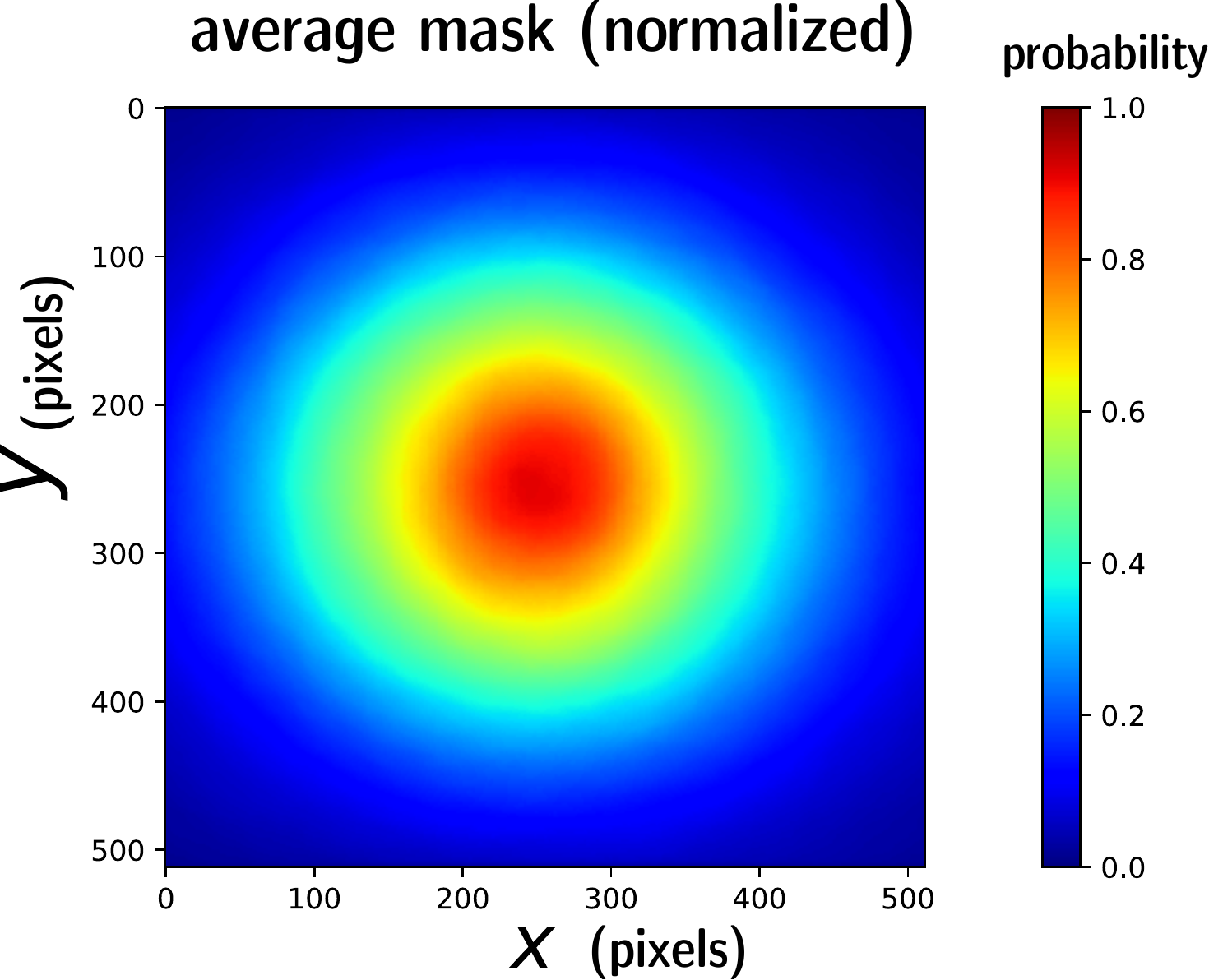} \quad
  \includegraphics[scale=.45]{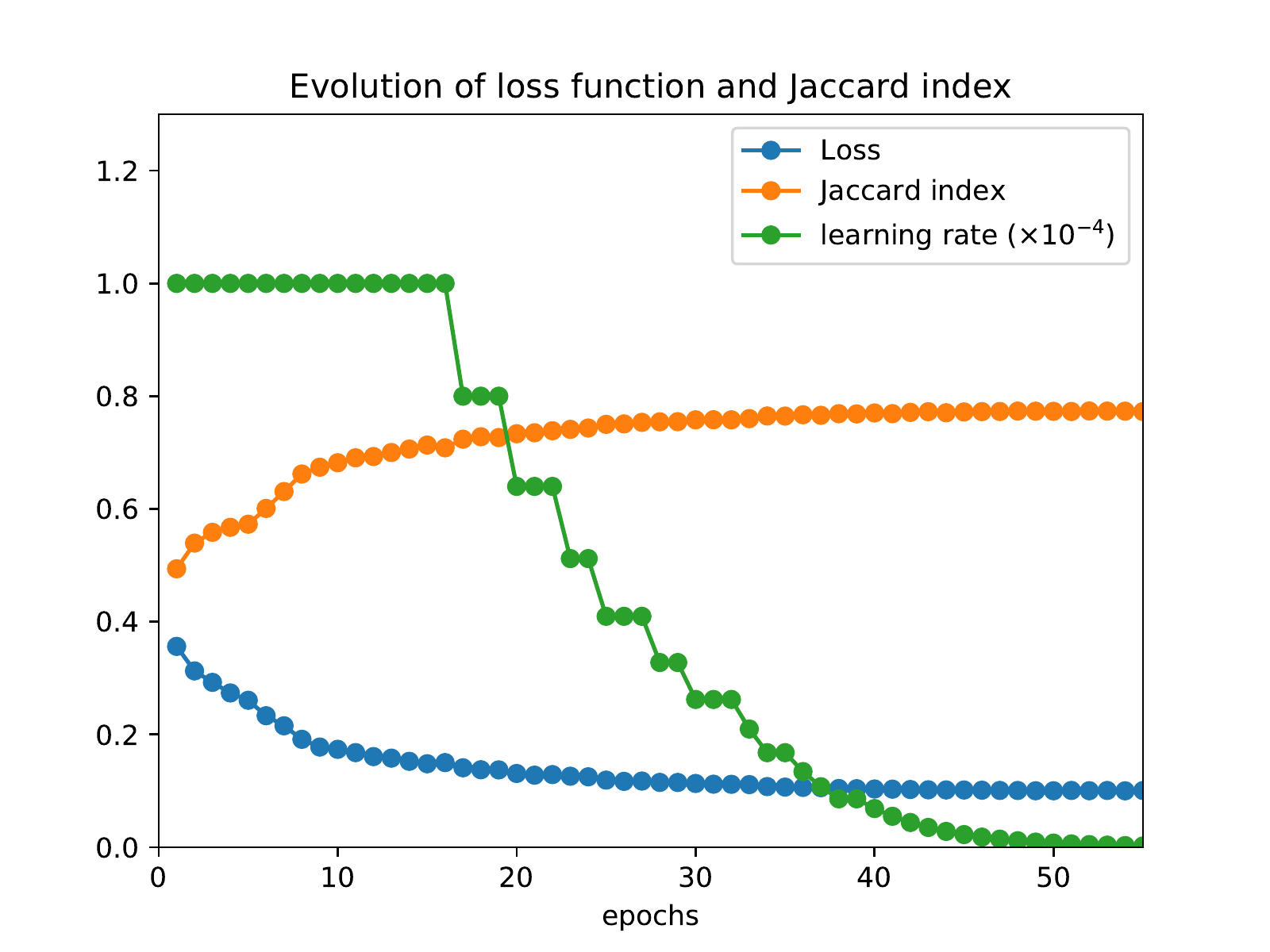}
  \caption{{\bf Left:} Probability that a pixel is part of a mole for the all training set. The probability at the center is around $p=.9$ and close to zero at the boundary ($p\approx .005$). {\bf Right}: Evolution of the loss function (blue) and Jaccard index (orange) per epochs.}
  \label{fig:average_mask}
\end{figure}

Before the training of the U-Net network \cite{ronneberger_u-net:_2015}, we first pre-processed the data-set. The very first statistic consisted in estimating the {\it average pixel} by estimating the proportion of each color ( $\langle Red, Green, Blue\rangle = (0.708, 0.582, 0.536))$ (red is slightly more prominent) and the standard deviation of the data-set $(0.0978,0.113,0.127)$. This information was used in the training to re-center the training set. Another important information was the average proportion of the mole in the image which was estimated as:
\begin{equation}
  \label{eq:propor}
  \text{prop. mole} = \frac{\#{ \text{pixels ground truth}}}{\#{ \text{pixels in image}}} =  0.214.
\end{equation}
Masks represent in average only $21\%$ of the image. The loss function was modified to even this ratio. Lastly, we also estimated the position of the mole inside the image. As expected, moles are more likely present in the center (see fig.~\ref{fig:average_mask}-left). This information can be used as a prior for a Bayesian approach (not used here).

\subsection{Training}

For the training of the U-net network, we used the cross entropy as loss function and the Adam optimizer to update the parameters. Learning rate was set-up as $10^{-4}$ and reduced when the loss function was not decaying. The evolution per epoch of a typical training is given in figure \ref{fig:average_mask}-right. The orange curve gives the evolution of the so-called Jaccard index used to rank the method. 
Usually data-augmentation methods were used (random flipping and image rotation). The code has been implemented using PyTorch.





\subsection{Post-processing}

\begin{figure}[ht]
  \centering
  \includegraphics[width=.47\textwidth]{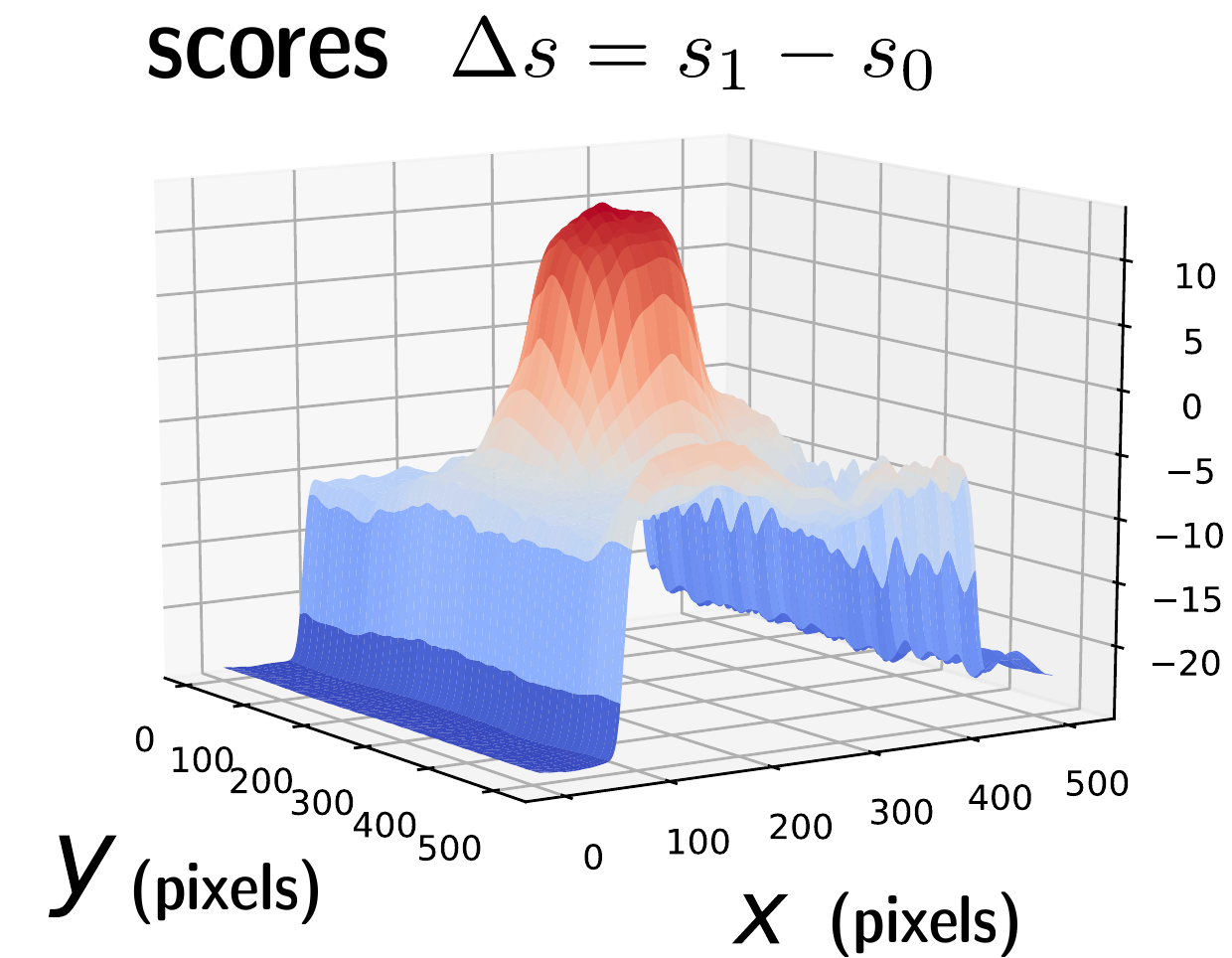} \quad
  \includegraphics[width=.47\textwidth]{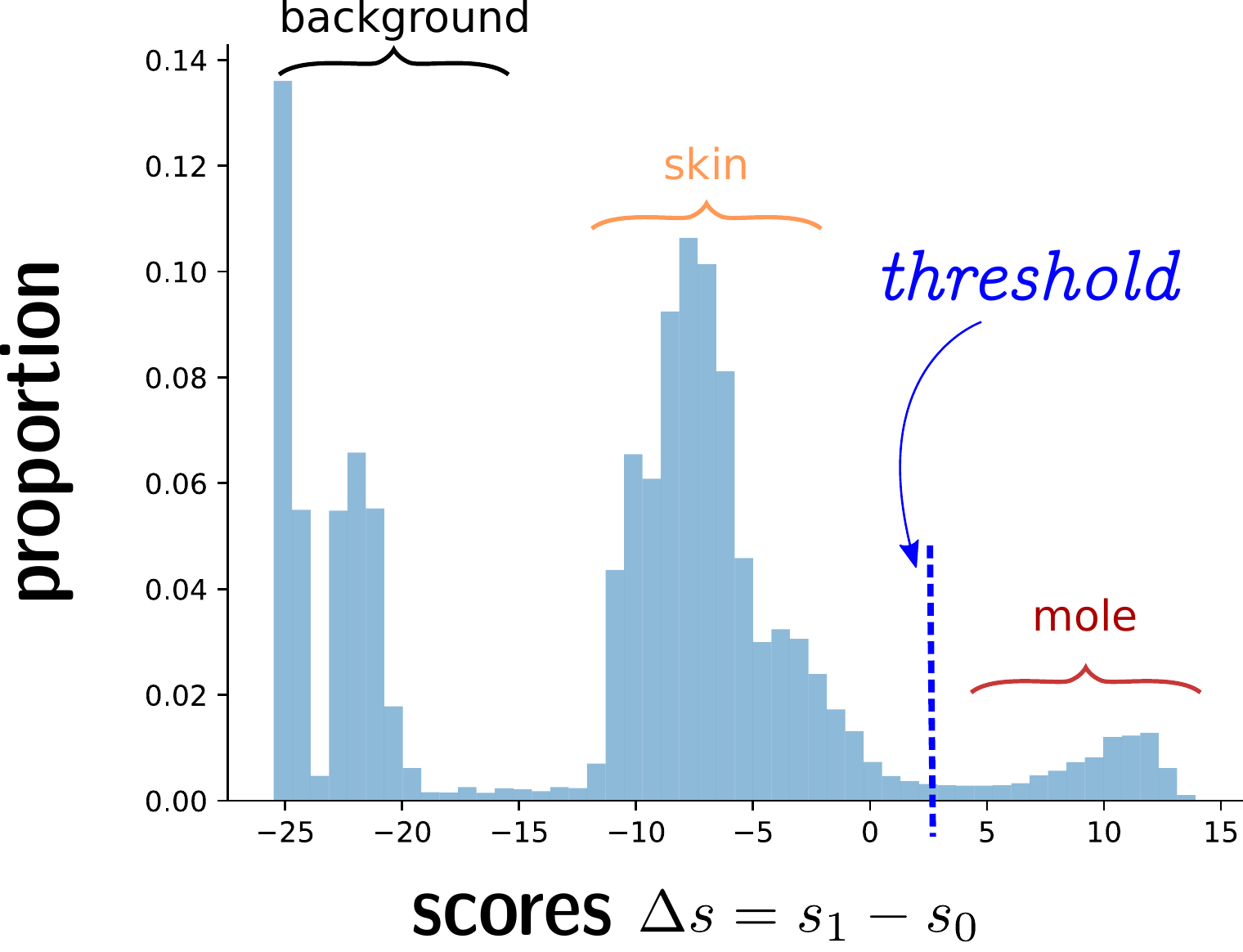}
  \caption{Distribution of the score (foreground - background, i.e. $\Delta s=s_1-s_0$) returns by the U-net network. Rather than taking a zero threshold to determine the mask, we first estimate the threshold separating distribution of scores (Otsu method).}
  \label{fig:post_process}
\end{figure}

The U-net gives a two \textit{scores} for each pixel (scores $s_0$ for the background and $s_1$ foreground). These scores $(s_0,\,s_1)$ are then transformed into probability $(p_0,\,p_1)$ via softmax function:
\begin{equation}
  \label{eq:score_proba}
  p_0 = \frac{\expo^{s_0}}{\expo^{s_0}+\expo^{s_1}},\quad  p_1 = \frac{\expo^{s_1}}{\expo^{s_0}+\expo^{s_1}}.
\end{equation}
The mask is then taken as the pixel with probability $p_1$ higher than $.5$. However, we find out that the masks can be further improved using several post-processing methods on the scores $(s_0,s_1)$. First, we apply a Gaussian filter on both scores ($\sigma=5$). Plotting the difference $\Delta s = s_1-s_0$ (see fig.~\ref{fig:post_process}-left), we observe three \textit{levels}: at the boundary the difference is below ($\Delta s<-15$), at the center where the mole is, the values are around $\Delta s>5$, and finally the \textit{normal} skin the values are in between $-10$ and $0$. The histogram of all these values $\Delta s$ (regardless of their position on the image) is given in fig.~\ref{fig:post_process}-right. The \textit{threshold} separating 'skin' to 'mole' seems to be around $\Delta s_*=2.5$. We use the Otsu algorithm \cite{otsu_threshold_1979} to estimate automatically this threshold.

\subsection{Results}

Our method score a Jaccard index of $.75$ over various test set. For the  validation test, the  network has a score of $.71$ due to the metric used (scores below $.65$ are considered zero).

\medskip

{\bf Acknowledgments} Authors would like to thank Thomas Laurent and Xavier Bressan for fruitful suggestions. S. Motsch thanks Renate Mittelman, Salil Malik and people at research computing (ASU) for their support during the computation.





\end{document}